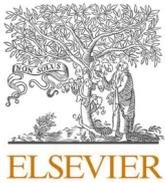
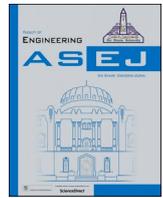

Full Length Article

# A novel facial recognition technique with focusing on masked faces

Dana A Abdullah [a,b,*], Dana Rasul Hamad [c], Ismail Y. Maolood [b,d], Hakem Beitollahi [c], Aso K. Ameen [a,b], Sirwan A. Aula [c], Abdulhady Abas Abdulla [e], Mohammed Y. Shakor [f,g], Sabat Salih Muhamad [c]

[a] Department of Computer Science, College of Science, Knowledge University, Erbil, Kurdistan Region, Iraq
[b] Department of Information System, ICTC Directorate, Ministry of Higher Education and Scientific Research, Erbil, Kurdistan Region, Iraq
[c] Computer Science Department, Faculty of Science, Soran University, Soran, Erbil, Kurdistan Region, Iraq
[d] Department of Information Technology, Technical College of Informatics, Sulaimani Polytechnic University, Sulaimani, Kurdistan Region, Iraq
[e] Artificial Intelligence and Innovation Centre, University of Kurdistan Hewlér, Erbil, Kurdistan Region, Iraq
[f] Information Technology Department, College of Computer and Information Technology, University of Garmian, Kalar, Kurdistan Region, Iraq
[g] Department of Computer, College of Science, University of Sulaimani, Sulaymaniyah, Kurdistan Region, Iraq

ABSTRACT

The recognition of the same faces masked and unmasked is a paramount function in preserving consistent recognition in public security, safety, and access control. Facial recognition technologies have been seriously tested with the widespread use of masks due to infectious diseases in recent years, which cover key facial areas and reduce identification levels. In this paper, we introduce a novel Masked-Unmasked Face Matching Model (MUFM) that uniquely leverages cosine similarity to match masked and unmasked face images a task that, to our knowledge, has not been addressed before. Our approach uses transfer learning with pre-trained VGG-16 for discriminative facial feature extraction followed by feature structuring using a K-Nearest Neighbors (K-NN) classifier. The most significant innovation is the utilization of cosine similarity to compare feature embeddings, such that strong identification is possible even when critical facial regions are obscured. To establish the model proposed, we have developed a comprehensive dataset from three different sources i.e., real-world pictures resulting in 95% recognition. This work not only addresses a vital gap in occluded face recognition but also offers a scalable solution to security and surveillance activities across environments with varying occlusion rates.

## 1. Introduction

Facial recognition has become increasingly vital to a variety of industries, from security and law enforcement to access control. Extensive mask-wearing has been a particularly tough challenge, undermining the performance of existing systems. Facial recognition has advanced significantly in recent years, driven by advances in artificial intelligence. This issue has led to incredible development of face detection algorithms, which enables high accuracy in numerous applications [1]. Besides, facial recognition has also emerged as a critical part of several systems, including user interaction interfaces and security systems [2,3]. The success of these applications relies on two significant factors: accurate facial feature analysis and reliable detection.

Cosine similarity is a good metric for finding the similarity between two vectors in terms of the cosine of the angle between the two vectors [4]. If the value of the cosine is close to zero for facial recognition, then it means that the features of unmasked and masked faces are closely similar and hence are likely to belong to the same person. This method allows precise matching of facial images, with features being vectors extracted through deep learning techniques such as Convolutional Neural Networks (CNNs).

Visual Geometry Group (VGG) architecture is a fundamental structure in implementing CNNs for deep learning image classification. VGG-16 consists of 13 convolutional layers and 3 fully connected layers and employs small 3x3 receptive field filters. Notably, VGG-16 contains 118 million parameters, which consume enormous memory and computational resources [5]. The model employs Rectified Linear Unit (ReLU) as an activation function to introduce non-linearity. For vector classification to images, K-NN is also a commonly used and efficient method, widely employed in most research studies [6].

The intrinsic variability and complexity of image information led to significant challenges in facial recognition, particularly when dealing with masked faces. Despite the application of advanced face detection techniques and methods like Haar Cascade Classifiers (HCC), Histogram of Oriented Gradients (HOG), and Dlib's Face Detection, there were shortfalls in handling occlusion and partial visibility. Inadequate eye detection and especially in masked scenarios also contributed to the shortcomings. Also, the lack of massive datasets with different masked and unmasked images of the same individuals, together with mask coverages at different levels, contributed to varying recognition performance. Traditional algorithms struggled to handle masks' occlusions, especially when masks were placed asymmetrically or where significant facial features were occluded. This highlights the necessity for continued innovation in detection algorithms and development of high-quality, diverse datasets in order to overcome the ever-present challenge of masked face recognition.

This paper answers this main research question: Can cosine similarity, in combination with transfer learning and conventional machine learning techniques, effectively match unmasked and masked face images with a high accuracy? In fact, this study aims to extend the research on facial recognition, particularly in challenging masked scenarios, by developing a model that applies cosine similarity to achieve effective face matching. The objectives of this paper are:

1. To discuss the effectiveness of cosine similarity in detecting masked and unmasked faces, an area where prior research has shortcomings.
2. To explore and identify suitable methods and algorithms that enhance facial recognition using cosine similarity.
3. In order to use these techniques in practice, to develop an advanced system that could be applied to security, social networks, etc.

Finally, the main contributions are utilized in this study is highlighted as follows:

- **Innovative Methodology:** We present the initial application of cosine similarity to match masked and unmasked face images, providing a new answer to a challenging problem.
- **Robust Feature Extraction:** By employing the transfer learning of VGG-16, our model has the ability to extract discriminative facial features even if the face is partially occluded.
- **Efficient Classification:** Integration of K-NN makes facial embedding classification easy and interpretable.
- **Comprehensive Dataset:** We have acquired an image dataset of three diverse sources incorporating real-world data to render our model strong under mixed conditions.
- **High Performance:** Our experimental results indicate that our method performs 95 % recognition accuracy, highlighting the practical viability of our approach for real-world applications such as surveillance and access control.

The structure of the paper is organized as follows. Section II discusses the related work, Section III proposes our methodology using cosine similarity to detect masked and unmasked faces. Section IV presents the experimental study and evaluates the models including results and discussion part. Finally, Section V concludes the paper.

## 2. Related work

Facial recognition has seen significant advancements in recent years, but research specifically focused on masked face recognition using cosine similarity remains limited. Various studies have explored deep learning techniques to address this challenge. [7] utilized pre-trained Convolutional Neural Networks (CNNs) like VGG-16, ResNet-50, and AlexNet to extract features from the eye and forehead regions of masked faces. These models have demonstrated effectiveness in masked face recognition but may struggle with generalization across diverse mask types and facial variations. A superior deep learning model, ArcFace, utilizes the Additive Angular Margin loss function, which has shown improved performance in large-scale face recognition tasks. [8] proposed ArcFace to enhance face recognition by maximizing class discriminability with a clear geometric meaning. The authors noted that ArcFace is robust to label noise and introduced sub-center ArcFace, allowing multiple sub-centers per class to mitigate real-world noise challenges. Despite its effectiveness in discriminative feature embedding and identity-preserving face synthesis, ArcFace still faces methodological limitations.

[9] tested a generic face recognition model on masked face images without additional training. They employed face embeddings and classified them using K-NN, Support Vector Machine (SVM) with a One-vs-Rest strategy, and Mean Embeddings (ME) via Sk-learn Nearest Centroid implementation. Their approach enhanced recognition precision in masked-face conditions by 9.09 %, demonstrating the efficacy of classifiers without additional training. [10] explored facial recognition using traditional feature extraction methods such as HOG, Local Binary Patterns (LBP), Principal Component Analysis (PCA), Speeded-Up Robust Features (SURF), and HAAR-like features. They used the K-NN algorithm for classification and achieved the highest accuracy rates (above 85 %) using HOG.

Similarly, [11] investigated face recognition without masks, relying on structural and feature similarity measures, edge detection, and Gaussian noise. They compared three models: Feature-Based Structural Measure (FSM), Structural Similarity Index Measure (SSIM), and Feature Similarity Index Measure (FSIM). FSM outperformed SSIM and FSIM, achieving the highest accuracy.

The COVID-19 pandemic necessitated research into AI-based facial recognition for masked faces. [12] proposed a system using MobileNetV2 and OpenCV for mask detection, and FaceNet with a multilayer perceptron for facial recognition. Their model, trained on 13,359 images (52.9 % masked), achieved 99.65 % accuracy in mask detection and 99.96 % for non-masked face recognition. [13] also addressed masked face recognition, employing SD-MobileNetV2 for mask detection, Multi-Layer Optimization (MLO) for face detection, and Robust Principal Component Analysis (RPCA) for occlusion handling. Their approach, enhanced with Particle Swarm Optimization (PSO) for K-NN feature selection, achieved a 97 % recognition rate and demonstrated improved robustness to occlusion.

[14] developed a masked face recognition system during the COVID-19 pandemic. They synthesized masks for benchmarks like IJB-B, IJB-C, FG-Net, SCface, and MS-1MV2 and tested model robustness against variations in pose, illumination, and age. Their study utilized ResNet-100 for feature extraction and novel loss functions such as Center Loss and Marginal Loss Angular SoftMax. The results showed that models trained on synthetic masked face data outperformed human recognition on benchmark datasets.

The growing interest in masked face recognition has led researchers to explore different methodologies. Some studies have leveraged cosine similarity for face verification tasks, such as the xCos metric, which improves interpretability. However, its application to masked face recognition remains unexplored. Additionally, methods like CosFace and ArcFace, which introduce additive angular margins, enhance class separability but require extensive computational resources [8,15]. Several approaches integrate deep learning with classical machine learning. [16] demonstrated improved accuracy by combining deep-learning-based mask detection with classifiers like K-NN. While effective, such pipelines are often computationally intensive. Attention modules have also been incorporated to focus on unoccluded facial regions, enhancing feature extraction but increasing model complexity. Our research introduces the Masked-Unmasked Face Matching Model (MUFM), which addresses gaps in existing literature. We utilize VGG-16 for feature extraction to minimize training data requirements and computational costs. Instead of deep learning classifiers, we employ K-NN for classification, ensuring simplicity and interpretability. The use of cosine similarity for comparing masked and unmasked images enhances accuracy in partial occlusion scenarios. To overcome dataset limitations, we curated a specialized dataset from three sources, ensuring robust evaluation.

Table 1 displays different face recognition studies which analyze various technological approaches for masked face recognition. It includes deep learning approaches like VGG-16, ResNet-50, and AlexNet for feature extraction, as well as metric-based methods such as xCos, ArcFace, and CosFace for enhanced class separability.

Our approach offers a streamlined yet effective solution by integrating transfer learning, classical machine learning, and cosine similarity, ensuring robust recognition even with facial occlusions.

## 3. Methodology

In this section, the general and specific methodology of this work is presented through a series of successive paragraphs and figures. The first steps include data collection from different sources followed by the data loading and data preprocessing stage. Then feature extracting, classifying, and measuring similarity between vectors occur. Last, the model is trained and tested on all masked and unmasked images to assess the performance of such models.

Fig. 1, illustrates the methodology stages of the proposed model (MUFM) for masked and unmasked face recognition, encompassing data collection, preprocessing, feature extraction, classification, and implementation, leading to matched outputs. All the stages will be presented in the next sections.

So, with this study, we proposed a wide range of methods to improve masked and unmasked face recognition using VGG-16 for feature extraction, K-NN for classification, and the metric of cosine similarity for measuring similarity among extracted features in the forms of vectors. First, the primary facial features were detected from both masked and unmasked images through the pre-trained deep learning model called the VGG-16, which can identify difficult visual patterns. The images obtained from prepared dataset were then pre-processed to feed into the VGG-16 model while keeping most of the layers of the VGG-16 model frozen and tuning other layers. The extracted feature embeddings procedure was converted to vectors and K-NN classified these vectors into put probable similar neighbors into the same bucket. Lastly, cosine similarity was computed to find the relation between masked and unmasked images, which can allow comparing feature embedding for the same subject under both conditions. This methodology provides a sound paradigm for both masked and unmasked face recognition.

### 3.1. Data collection

One of the primary challenges encountered in this study was the limited availability of image datasets containing the same individuals with and without masks. This scarcity of data is a significant limitation, as datasets featuring identical faces both masked and unmasked are exceedingly rare or virtually nonexistent. While there are numerous data sets available for unmasked faces, finding corresponding masked images of the same individuals is particularly challenging. To address this issue, we made extensive efforts to search and gather images from various sources. After a thorough and resource-intensive search, we successfully compiled a dataset of 858 images that meet the study's criteria. This dataset includes paired images of the same individuals, both with and without masks, enabling us to proceed with our analysis despite the data constraints. These data have been collected from three different sources such as [20] celab-A-HQ from the Kaggle website as it includes several images each image taken in different directions. The second part has been collected from the Zenodo Website which is sourced in [21]. Other images were real data collected by the current research authors. Additionally, both male and female images are available in the dataset as shown in Fig. 2, 58 % of the images for male, and 42 % of images are female. And, the ages range from 21 to 75 years, notably 65 % of the images feature individuals aged between 21 and 50. Also, the dataset contains different images with faces covered with different types of masks and ensures the model is capable of accurately detecting masked faces regardless of the mask types.

Fig. 3 Shows the process of collecting data from the above-mentioned sources, based on this process firstly the image data was directly taken by the authors in order to be utilized as real data, however, the real collected data was not enough to provide a promising result that is why the other sources were utilized to collect data like the previously mentioned sources then these data finally became a single and integrated dataset for training and testing phases.

The percentage of taken images from different sources has been shown in Fig. 4 That displays there 60 % of the image dataset has been collected from the Zenodo Website, and 25 % of the data has been collected from Kaggle referenced in [22] and 15 % of the image dataset was real data particularly collected by the authors.

After collecting the required data and combining them into one single dataset, then they have to be pre-processed as discussed in the next section.

After collecting the image dataset, we proceeded with an extensive image preprocessing stage to ensure the data was well-prepared for model training. The pre-processing steps included the following:

1. Folder Organization: We created two separate folders: There should be one folder for people without masks, and the other one for people with masks. This kind of clear segregation is important in order to properly label and in further activities during the model training stages.
2. Image Resizing: To keep the scale consistent, all the images available in both folders were altered to fit the standard size. It is very important to do it because otherwise different images to be used in the process take different sizes, and it will mislead the model and slow down the training process.

**Table 1**
Summary of key studies in masked face recognition.

| Study | Techniques/Methods | Algorithms Used | Results |
|---|---|---|---|
| Masked Face Recognition Using Deep Learning Models [17] | Pre-trained CNNs (VGG-16, ResNet-50, AlexNet) for feature extraction from eye and forehead regions | VGG-16, ResNet-50, AlexNet | Effective feature extraction from visible regions of masked faces |
| xCos: An Explainable Cosine Metric for Face Verification [18] | Explainable cosine similarity metric for face verification | xCos Metric | Enhanced interpretability in face verification tasks |
| ArcFace: Additive Angular Margin Loss for Deep Face Recognition [8] | Additive angular margin loss function to improve class separability | ArcFace | Improved performance in large-scale face recognition tasks |
| CosFace: Large Margin Cosine Loss for Deep Face Recognition [15] | Reformulated softmax loss as cosine loss with margin to enhance discrimination | CosFace | Achieved state-of-the-art performance on multiple face recognition benchmarks |
| MFCosface: A Masked-Face Recognition Algorithm Based on Large Margin Cosine Loss [19] | Masked-face image generation algorithm; large margin cosine loss; attention mechanism | MFCosface, Attention Mechanism | Improved accuracy in masked-face recognition |

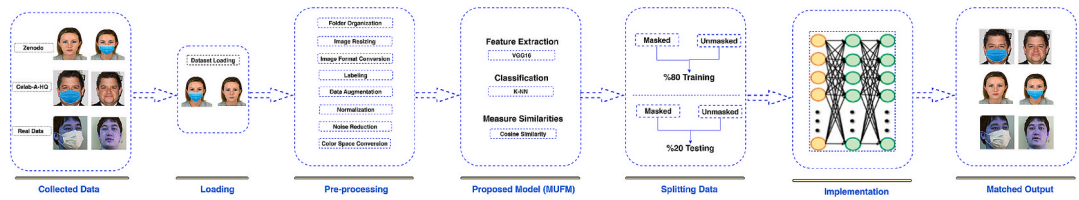

**Fig. 1.** Methodology study.

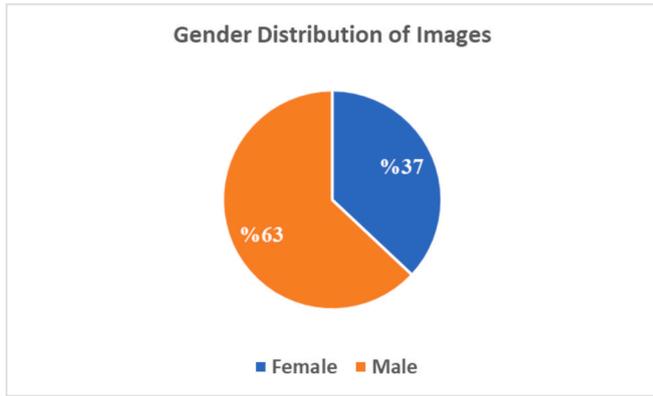

**Fig. 2.** Gender distribution of dataset images.

3. Image Format Conversion: Due to the flexibility of the dataset that contained different file formats of images such as JPEG, BMP, GIF, and many others, convert all images to PNG format. This brings consistency in the quality of images and does away with the problems that would be encountered during training in different file formats.
4. Labeling: Every image in the dataset was carefully annotated according to the given category. In the "without masks" set of images, the images were labelled as such, whereas images in the "with masks" set were given labels that correspond to them. This labelling is required in especially in supervised learning wherein the model requires proper labelling of data for training.
5. Data Augmentation: To improve the dataset's quality and quantity, it was decided to apply several steps of data augmentation. These included different operations like rotation, flip, zooming, and shifting which help in creating more variations artificially. There is one thing that augmentation is significant in making, which is avoiding overfitting and enhancing the generality of the model.
6. Normalization: We normalized the pixel values of all images to bring them to a common scale, typically between 0 and 1. This step helps in speeding up the convergence of the model during training and ensures that the model performs optimally.
7. Noise Reduction: Techniques like Gaussian filtering were applied to reduce noise in the images. This process helps in improving the quality of the images, making it easier for the model to detect relevant features during training.
8. Color Space Conversion: Depending on the specific requirements of the model, we converted images from RGB to grayscale, where necessary. This conversion can reduce the computational load and simplify the model if color information is not critical for the task.

The above stages are presented in Fig. 5.

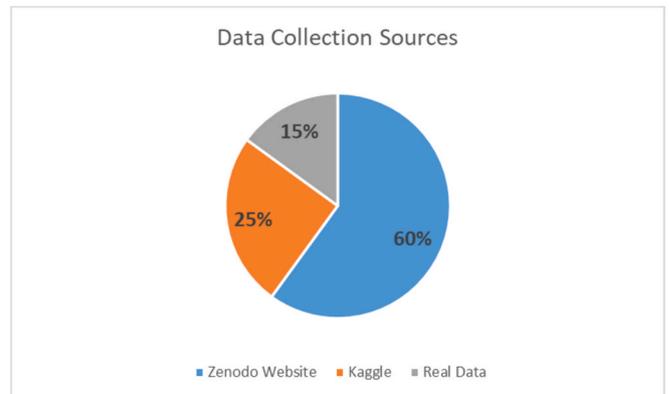

**Fig. 4.** Dataset sources rates.

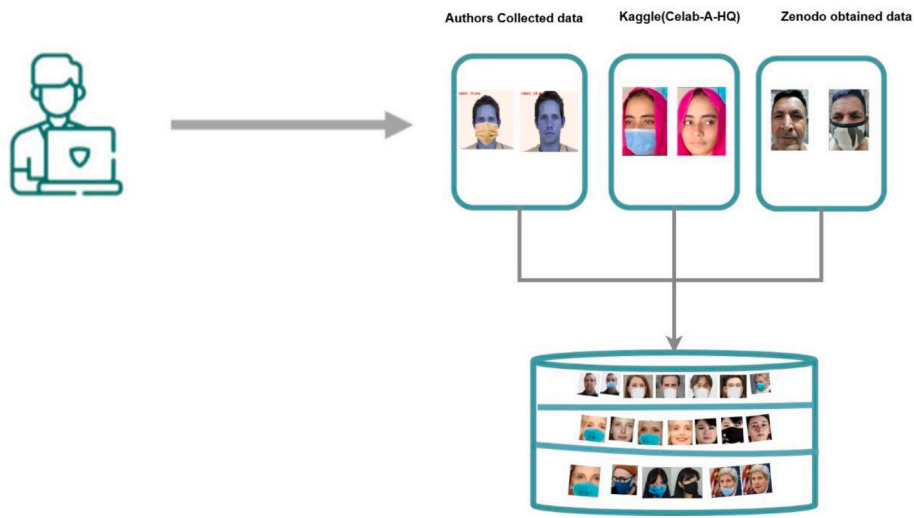

**Fig. 3.** Data collection process.

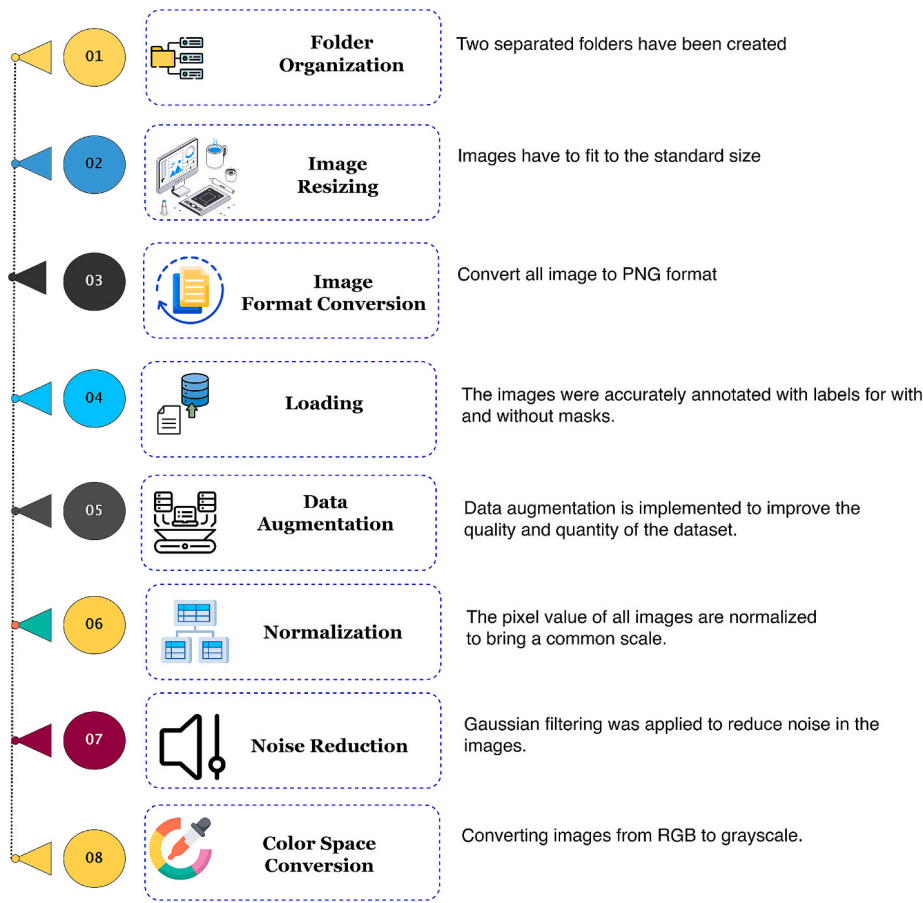

Fig. 5. Data pre-processing stages.

### 3.2. Proposed algorithms and techniques

To obtain the required results there are three steps (feature extraction, classification, and measure similarity) in the (MUFM) model that have been used, and they have also been implemented with the same model. So, in this section, these three techniques will be stated, however, the brief overview of the development process using these three techniques is shown in Fig. 6.

### 3.3. Convolutional neural network

In this research, the first crucial step is to extract important features from both masked and unmasked images of the same individual. To achieve this, a CNN is utilized, which is typically designed with a series of layers built specifically to capture and process visual patterns [23]. While general CNN architecture, consisting of standard layers, might perform adequately in some cases, it was found to be insufficient for this task of distinguishing features between masked and unmasked faces. The general CNN failed to capture the subtle variations and essential features needed for accurate recognition of the same person in both scenarios.

To avoid this peculiarity, it was decided to use a more complex, deep learning-based model for feature extraction. In such context, VGG-16, which is a readily available deep architecture pre-trained CNN model has been chosen. VGG-16 is specifically suitable for this purpose because of its capability to learn the intricate feature hierarchies and generate comparatively better discriminant embedding. This model enables the better extraction of specific facial features for a subject in both the masked and non-masked instances, or the general identification of the same subject in both these settings [24].

#### 3.3.1. Visual geometry group VGG-16

VGG-16 is a method in CNN and was developed by the VGG at Oxford University in the year 2014. This has 16 layers, 13 of which are convolutional layers and three of which are fully connected layers; it is popular for images classification and feature extraction since its design is basic but effective. VGG-16 makes uses of small filters of 3x3, ReLU activation functions, and max-pooling layers to obtain a pyramid of features ranging from simple edges to complex textures. Originally trained on millions of images of objects and animals from the ImageNet dataset, it is typically employed via transfer learning for other tasks such as the current forgery detection for extracting features in cases like face with mask and face without mask the current forgery detection. Role of VGG-16 in Masked and Unmasked Face Recognition.

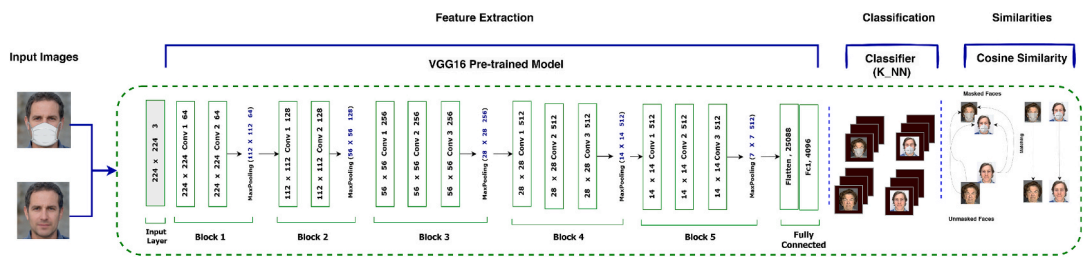

Fig. 6. MUFM processing model.

In the case of both mask-wearing and no mask, the VGG-16 model is a feature extractor, that identifies only the visible part of the face (the jaw and eyes behind the mask and the forehead when the face is uncovered), while simultaneously training a network that learns to extract representations invariant to masking. Thus, it captures outstanding features that are edges, textures, and shapes, which enable the model to learn and generalize despite occlusion.

*3.3.2. Techniques for feature extraction*

Pre-processing: Input images are resized to 224x224 pixels and normalized to ensure consistent pixel values before being fed into VGG-16.

Transfer Learning and Fine-Tuning: VGG-16's lower layers (which teach basic features) are frozen, while the upper layers are fine-tuned on a dataset of masked and unmasked faces, adapting the model to the task.

Feature Embeddings: Both masked and unmasked images are converted into feature vectors (embeddings). These vectors are then compared using metrics like cosine similarity to determine if the images belong to the same person.

VGG-16 is used as a pre-trained model for feature extraction purposes, and the convolutional base of VGG-16 was kept, the fully connected layers are replaced. When the convolutional layer has ended the "Global Average Pooling (GAP)" was utilized to convert the feature maps into a single vector. The global features of the image were represented based on this vector, this will offer of reducing the dimensionality and making it easier to handle.

*3.4. K-Nearest neighbors (K-NN)*

K-NN is a non-parametric algorithm used for classification, which works by finding the closest K data points (neighbors) to a given input based on a chosen distance metric, like Euclidean distance. In this research, K-NN is applied after VGG-16 has extracted features from both masked and unmasked images of the same individual. These features are transformed into vectors, representing the key facial characteristics.

The role of K-NN here is to group and classify these feature vectors, identifying the nearest neighbors in the embedding space. Once K-NN has separated these extracted features, they are ready for comparison using cosine similarity, which measures the similarity between the masked and unmasked face embeddings to determine if they belong to the same person.

*3.5. Cosine similarity*

Cosine similarity is a metric used to measure the similarity between two vectors by calculating the cosine of the angle (θ) between them. Mathematically, it is represented as the dot product of two vectors, A and B, divided by the product of their magnitudes, as shown in Equation 1.

$$Cosine\ Similarity = \cos(\theta) = \frac{A \cdot B}{\|A\|\|B\|} \quad (1)$$

[25].

Here, A. B is the dot product of the two vectors, the graphics A and B, and ‖A‖ & ‖B‖ stands for their norms. The cosine similarity value is from −1 up to 1. Statistics showing that the value is equal to 1 mean that the vectors are similar, that is they have a positive orientation, while the value is equal to 0 implies that the vectors are perpendicular, that is they are not related and the angle between them is 90 degrees. Cosine similarity of −1 means vectors is at a subterranean level of dissimilarity with the angle of 180, this is the threshold of the model.

In the use of face recognition, therefore, a cosine similarity close to 1 means that the two vectors (faces) are closely related and should represent the same face. On the other end of the scale, S < 0 means that the two faces are most different, in other words, the faces are orthogonal. This is why cosine similarity is helpful in areas such as machine learning and pattern recognition, as amply demonstrated in the word embedding problem above, it is based on the direction rather than the magnitude of the data. For this reason, it is a very effective measure for comparing the feature vectors and for applications where the relative direction of data is important as compared to the direction magnitude.

In this study, the features from masked and unmasked images of the same person, as pre-processed by VGG-16, are converted to vectors. They then used the K-NN method to group and categorize these vectors. Cosine similarity has its work to do after K-NN has partitioned the features. They are used to quantify the level of similarity between the vectors which represent masked and unmasked face. Higher the value of cosine similarity, the match of features extracted from two images implies that the system can identify the subject whether with or without a mask. Cosine similarity provides a better platform to find out the similarity of the facial structure or features under various conditions as shown in Fig. 7, how the cosine similarity works for input image data and apply feature extraction technique.

## 4. Experimental study

The implementation of this report employs Feature extraction using the VGG-16 model and K-NN in classifying the images of faces with and without masks. The code first links Google Drive to be able to use the dataset downloaded, then loads and pre-processes the images, including using data augmentation to add some level of randomness. Based on VGG-16, a personalized model is developed and then pre-trained and trained for the images' classification. Next, embeddings are obtained from the model and the k-neighbor nearest neighbors' classifier is applied on the images. Cosine similarity is then used as a method of detecting and pairing masked and unmasked images given their embeddings. Lastly, the code brings up two pictures with and without the mask side by side if the model finds them similar, to show its matching function.

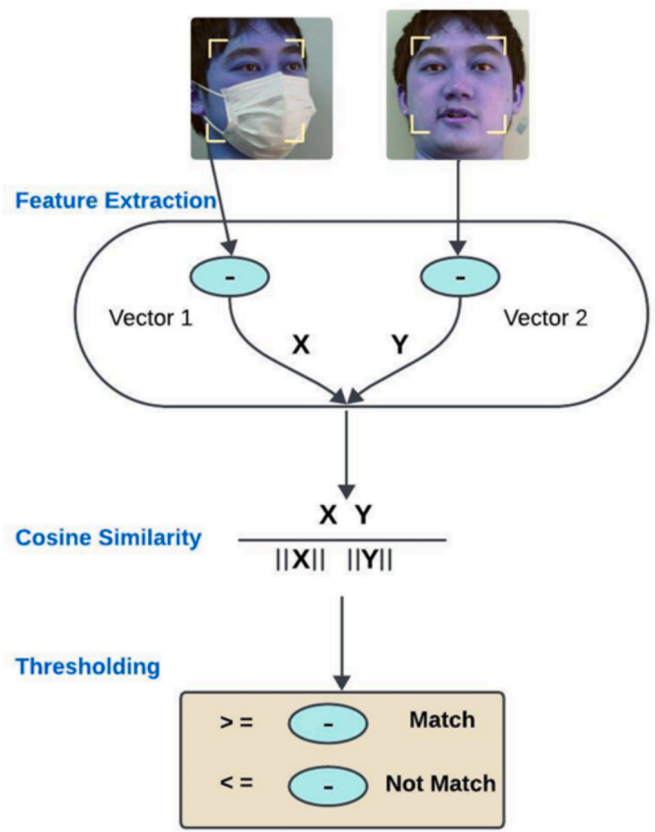

**Fig. 7.** Feature extraction and cosine similarity for input image data.

Python programming was performed in the Google Colab environment in the context of a practical experiment, which utilizes its advanced hardware notably a GPU which significantly improves computational efficiency and speeds up the process of training phases of the model. A total of 20 epochs were conducted for the training process. These iterations are chosen with care to balance computational resources with desired model performance enabling the network to earn effectively without overfitting.

*4.1. Results*

The model is successfully able to put into operation a deep learning model using the VGG-16 structure for identifying the same faces with and without the masks. To improve the model's training process initially, dense layers are added, and it has been used to increase the resistance of data augmented. The model loses accuracy only slightly when determining whether faces are masked or not.

Moreover, embeddings extracted from the model are utilized by a K-NN classifier, which also performs well, accurately distinguishing between faces with and without masks. The use of cosine similarity to match image pairs further illustrates the model's ability to identify the same individuals regardless of mask usage, demonstrating its effectiveness in face recognition tasks involving masks.

Based on the implementation of the current work there are comprehensive results achieved; the majority of the epochs obtained over %90 or val_accuracy with every 20 epochs the average result is %95 val_accuracy. The direct faces which are shown in Fig. 8 between masked and unmasked images were detected properly.

Even for non-direct faces which is illustrate in Fig. 9, was recognized which proves the effectiveness of the current work technique.

An analysis of various masked face recognition methods appears in Table 2 which describes their performance through different computational methodologies. The current research using cosine similarity achieved a 95 % accuracy level that surpassed traditional machine learning and feature extraction approaches for dealing with facial occlusions. Traditional machine learning models such as K-NN (85 %), Logistic Regression (78 %), and LDA (72 %) demonstrated competitive results but struggled with occluded facial features. Feature-based methods like HOG (85 %) and FSM (86.61 %) performed well, while techniques such as SSIM (57.73 %) and FSIM (28.03 %) were significantly hindered by occlusions. Compared to these approaches, our deep learning-based method, which integrates cosine similarity and K-NN classification, ensures superior face matching. This makes it a more reliable solution for real-world applications where recognizing individuals with and without masks is essential. The results confirm that our method provides the most robust and scalable solution for masked face recognition.

Fig. 10 shows the study curve which represents the training accuracy and validation loss of 20 epochs, displays how well the model is fitting the training data, and rapidly decreases loss within the first few epochs and indicates the model is quickly learning from the training. The red line represents the validation loss at the beginning it decreases rapidly. However, it stabilizes around the third epoch. This specifies that the model has learned the overall patterns in the data. Both green and yellow lines track the accuracy of the model on training and validation data. They increase quickly and achieve 100 % by around epoch 4. This suggests that the model strongly fits the training and performance of the validation data.

*4.2. Discussion*

This code can identify the same faces with or without the influence of a mask, having adopted transfer learning from a VGG-16 model. Using the code thus trained on the VGG-16 algorithm, the current implementation recognizes the difference between a face that has been masked and a face that has been covered up. Freezing the base layers of VGG-16 and adding some more additional dense layers also prevents the model from overfitting, which is mostly important if provided a large and complex dataset.

The ability of the model to generalize has been premised by the high accuracy recorded on both the training and the validation set. Among these, data augmentation is presented through the image data generator class and is critical to generalization, as it adds variability in the training dataset thus preventing overfitting due to facial conditions.

Apart from the deep learning model, there is the application of a K-NN classifier developed on embedding files from the VGG-16 model,

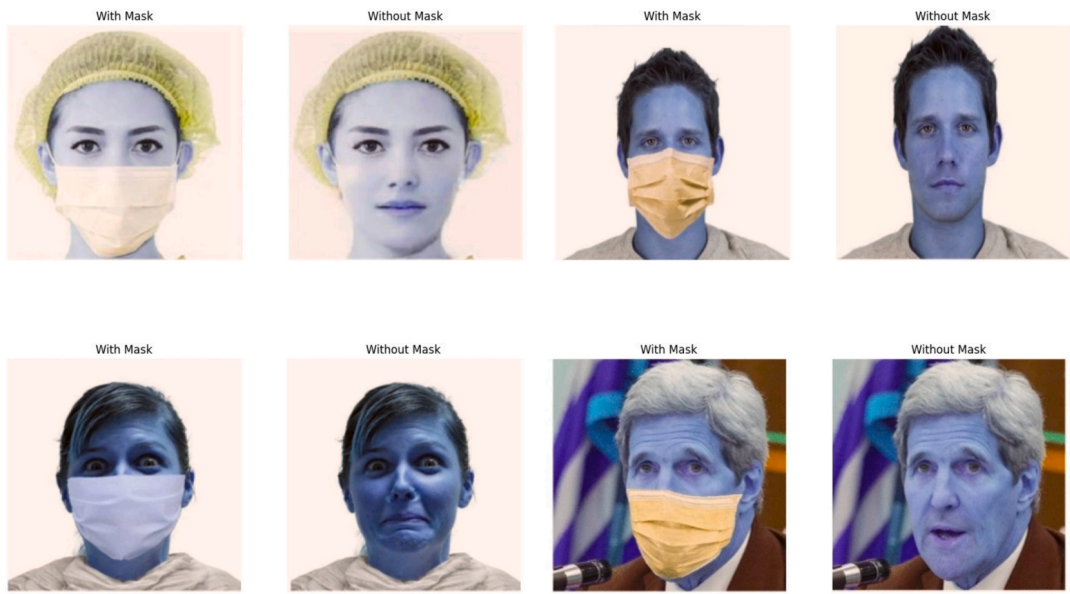

**Fig. 8.** Direct face images with and without mask.

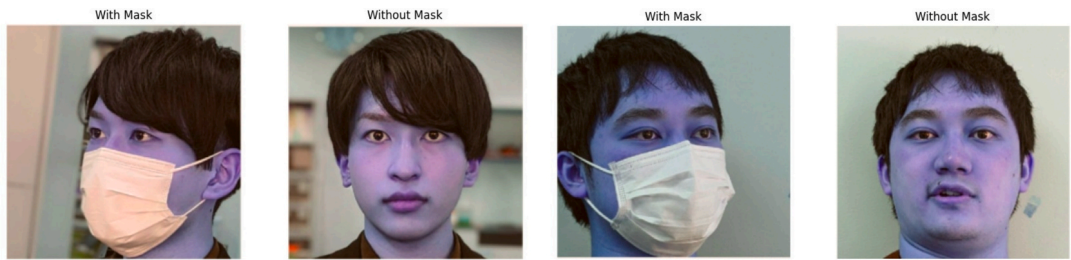

**Fig. 9.** Non-direct face images with and without mask.

**Table 2**
Performance comparison of masked face recognition techniques across different studies.

| Study | Used techniques | Results |
| --- | --- | --- |
| Current Study | Cosine Similarity | 95 % |
| [11] | SSIM | 0.5773 |
|  | FSM | 0.8661 |
|  | FSIM | 0.2803 |
| [26] | SVC | 70 % |
|  | LDA | 72 % |
|  | K-NN | 46 % |
|  | DT | 37 % |
|  | LR | 78 % |
|  | NB | 65 % |
| [10] | HOG | 85.0 % |
|  | LBP | 82.5 % |
|  | HOG & LBP | 82.5 % |
|  | Harris | 77.5 % |
|  | Surf | 55.0 % |
|  | PCA | 72.5 % |
|  | K-NN | 85 % |

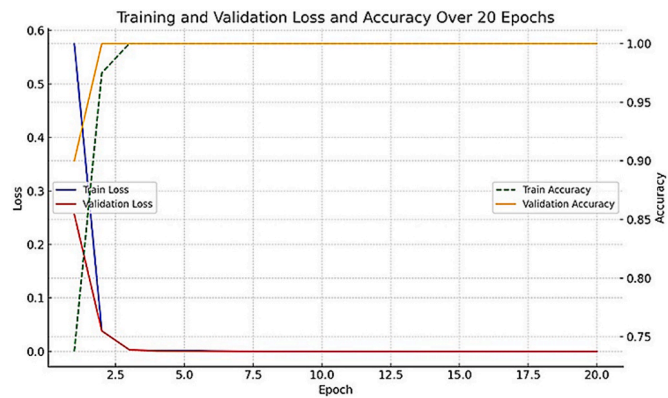

**Fig. 10.** Model training and validation loss and accuracy.

which forms the hybrid approach. The ability of the K-NN classifier, which achieved perfect accuracy at the test set level, can be attributed to the discriminative characteristic of the learned embedding, in which the current model is turning out to be quite effective in discriminating between the same faces with and without the masks.

Visualization of matched pairs of images (with and without masks) based on cosine similarity also shows a clear picture about the performance of the proposed model. This visualization does more than just ratify the model's proficiency; it is also beneficial for identifying how the changes in recognizing the same faces with masked and unmasked function optimally.

Although the results obtained with the proposed model are promising, they must be evaluated again on a large dataset with a high level of variability in order to validate the reliability of the approach to real-world conditions. Moreover, it can be also beneficial to apply the model in conditions when it will be trained on new data from time to time (for example, using online learning) in order to improve its performance in conditions when the data changes its parameters from time to time.

## 5. Conclusion and future work

This work proposed the Masked-Unmasked Face Matching Model (MUFM), a new method specifically created to solve the essential problem of matching faces between masked and unmasked conditions. By employing cosine similarity on transfer-learned VGG-16 network feature embeddings structured through K-NN uniquely, MUFM is able to perform robust recognition even when major facial features are covered. The creation of a heterogeneous dataset from real images permitted the model to be trained and tested with a 95 % recognition rate. This degree of performance reflects MUFM's potential to significantly enhance security and surveillance in settings with varying occlusion rates, thereby filling a significant gap in current face recognition technology.

Future work can extend MUFM's capabilities in a number of directions. First, pushing attention mechanisms out to feature extraction would probably further increase the selectivity of the model to discriminative facial areas. Second, investigating how much the model generalizes when facing more extreme illumination and viewpoint changes, or when there are other types of occlusions varying across the face besides masks, would be worthwhile. Lastly, creating an actual implementation of MUFM for real-world applications in security systems and access control would be a major breakthrough. Moreover, investigating the ethical aspects of applying masked face recognition in various situations is extremely crucial and should be taken into account in future research.

### CRediT authorship contribution statement

**Dana A Abdullah:** Writing – review & editing. **Dana Rasul Hamad:** Writing – original draft. **Ismail Y. Maolood:** Software. **Hakem Beitollahi:** Supervision. **Aso K. Ameen:** Visualization. **Sirwan A. Aula:** Methodology. **Abdulhady Abas Abdulla:** Validation. **Mohammed Y. Shakor:** Formal analysis. **Sabat Salih Muhamad:** Visualization.

### Declaration of competing interest

The authors declare that they have no known competing financial interests or personal relationships that could have appeared to influence the work reported in this paper.

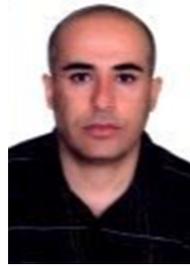

**Dana Ali Abdullah**, I currently work full time as an Employee in the Information Systems Department, ICTC Directorate of the Ministry of Higher Education and Scientific Research in the Kurdistan Region and I work part time as Assistant Teacher from various private universities in KRD. I earned my MSc degree in Computer Networking from the Department of Engineering at Sheffield Hallam University in the UK. Additionally, I obtained my BSc degree in Computer Science from Salahaddin University-Erbil in Iraq. My research interests which include AI-Based Networking, Telecommunication and Mobile, IoT (Artificial Intelligence (AI) Integrated Internet of Things (IoT), NLP, and Robotic and AI. GSC: https://scholar.google.com/citations?hl=en&user=w60L0LgAAAAJ ORCID: https://orcid.org/0009-0009-7610-3157

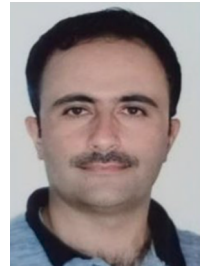

**Dana Rasul Hamad** is a skilled academic and Assistant Lecturer in the Computer Science Department at Soran University. Holding an MSc in Databases from the UK, Dana specializes in artificial intelligence, algorithm hybridization, and optimization, with a focus on health disease diagnostics. Their innovative work bridges theoretical knowledge with practical applications. Has developed diverse technological solutions, including mobile apps, websites, and Windows-based applications, showcasing technical versatility. With expertise in multimedia, excel as a graphic designer, enhancing text, images, videos, audio, and animations. Served as the Head of the Computer Science Department for 1 year and 8 months, driving academic progress. With extensive teaching experience across multiple institutions, Dana inspires future computer scientists, actively participate in international conferences, contributing to discussions on emerging technologies. Dedicated to teaching and research, Dana continues to advance the fields of computer science and artificial intelligence through innovation and excellence.

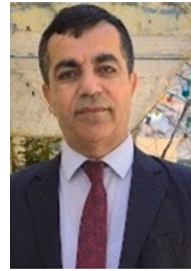

**Ismail Yaqub Maolood** I am currently Director of Statistics and Planning in the Ministry of Higher Education and Scientific Research, Erbil, Kurdistan Region, Iraq. I received my Ph.D. degree in the School of Computer Science and Technology at the Huazhong University of Science and Technology (HUST) in China, my M.Sc. degree in computer science from the Universiti Teknologi Malaysia in Malaysia, and a B.Sc. degree in computer science from Salahaddin University-Erbil. My research interests include the IoT, image processing, networking, and cloud computing. GSC: https://scholar.google.com/citations?hl=en&user=tdB8KHQAAAAJ ORCID: https://orcid.org/0000-0003-1683-1493

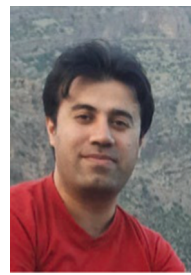

**Hakem Beitollahi** received his B.S. degree in computer engineering from University of Tehran, in 2002, the M.S. degree from Sharif University of Technology, Tehran, Iran, in 2005, and the Ph.D. degree from Katholieke Universiteit Leuven, Belgium, in 2012. He started working in School of Computer Engineering, Iran University of Science and Technology (IUST) as Assistant Professor and the Head of the Hardware and Computer Systems Architecture Branch. He left IUST at 2023. He is currently an assistant professor at department of computer science, Soran University. His research interests include hardware accelerator for artificial intelligence (AI), AI domain including machine/deep learning and hardware/network security.

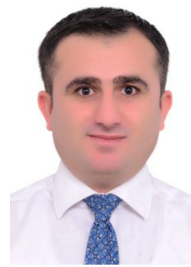

**Aso Khaleel Ameen**. I work full time as a head in the Information Systems Department, ICTC Directorate of the Ministry of Higher Education and Scientific Research in the Kurdistan Region, and part time as a lecturer at various public and private universities in KRI. I earned my MSc degree in Software Engineering from Firat University in Elazig, Turkey. Additionally, I obtained my BSc degree in computer science from Salahaddin University-Erbil in Iraq. My research interests are machine learning, data science, social network analysis, and natural language processing. GSC: https://scholar.google.com/citations?hl=en&user=3tL2kesAAAAJ ORCID: https://orcid.org/0000-0002-7037-7546

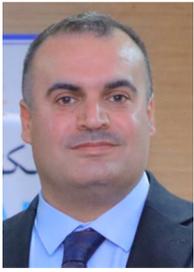

**Sirwan A. Aula** is an Assistant Lecturer at Soran University, specializing in Computer Science and Database Systems. With over a decade of academic experience, he has contributed to teaching, administration, and research in various capacities, including as vice dean of the Faculty of Science. His expertise includes programming, database management, and optimization research.

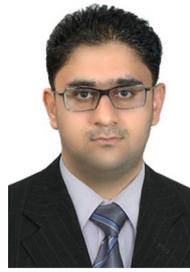

**Mohammed Y. Shakor** received the Master of Science degree from the Computer Science Department, College of Science, University of Sulaimani, in 2019. He is currently a Lecturer with the University of Garmian. He is also an accomplished academic professional with a profound expertise in computer science. During this tenure, he exhibited a remarkable aptitude for advanced concepts and demonstrated a keen interest in cutting-edge developments within the field. His research interests include cloud security, cryptography, deep learning, and cloud computing. He has developed innovative methods and techniques to enhance accuracy and efficiency in these fields.

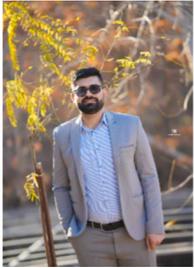

**Abdulhady Abas Abdulla**, born on October 15, 1997. I have a master's degree in computer science with an emphasis on NLP. Currently As research at the UKH Centre. My main interests are in Artificial Intelligence, Deep Learning, Natural Language Processing, and Speech Recognition.